\documentclass{article}
\usepackage{ijcai22}

\usepackage{times}
\usepackage{soul}
\usepackage{url}
\usepackage[hidelinks]{hyperref}
\usepackage[utf8]{inputenc}
\usepackage[small]{caption}
\usepackage{graphicx}
\usepackage{amsmath}
\usepackage{amsthm}
\usepackage{booktabs}
\usepackage{algorithm}

\usepackage{diagbox}
\usepackage{multirow}
\usepackage{arydshln}
\usepackage[noend]{algpseudocode}
\usepackage{amssymb}
\usepackage[switch]{lineno}
\usepackage[dvipsnames]{xcolor}
\usepackage{algorithmicx,algorithm}

\title{Comparison Knowledge Translation for Generalizable Image Classification}

\author{
    Anonymous IJCAI submission
    \affiliations
    Paper ID 2069
}

\author{
Zunlei Feng$^{1,4,5}$\and
Tian Qiu$^1$\and
Sai Wu$^{1}$\and
Xiaotuan Jin$^{3}$\and
Zengliang He$^{3}$\and
Mingli Song$^{1,4,5}$\and
Huiqiong Wang$^{2}$\footnotemark[1]
\affiliations
$^1$Zhejiang University\\
$^2$Ningbo Research Institute, Zhejiang University\\
$^3$Hangzhou Honghua Digital Technology Co., Ltd.\\
$^4$Shanghai Institute for Advanced Study of Zhejiang University\\
$^5$Alibaba-Zhejiang University Joint Research Institute of Frontier Technologies\\
\emails
\{zunleifeng,huiqiong\_wang\}@zju.edu.cn
}
\begin{document}

% \linenumbers

\maketitle

 \def\mathbi#1{\textbf{\em #1}}

\renewcommand{\thefootnote}{\fnsymbol{footnote}}
\footnotetext[1]{Corresponding authors.}

\begin{abstract}

Deep learning has recently achieved remarkable performance in image classification tasks, which depends heavily on massive annotation. However, the classification mechanism of existing deep learning models seems to contrast to humans' recognition mechanism. With only a glance at an image of the object even unknown type, humans can quickly and precisely find other same category objects from massive images, which benefits from daily recognition of various objects. In this paper, we attempt to build a generalizable framework that emulates the humans' recognition mechanism in the image classification task, hoping to improve the classification performance on unseen categories with the support of annotations of other categories. Specifically, we investigate a new task termed Comparison Knowledge Translation (CKT). Given a set of fully labeled categories, CKT aims to translate the comparison knowledge learned from the labeled categories to a set of novel categories. To this end, we put forward a Comparison Classification Translation Network (CCT-Net), which comprises a comparison classifier and a matching discriminator. The comparison classifier is devised to classify whether two images belong to the same category or not, while the matching discriminator works together in an adversarial manner to ensure whether classified results match the truth. Exhaustive experiments show that CCT-Net achieves surprising generalization ability on unseen categories and SOTA performance on target categories.
\end{abstract}

 \def\mathbi#1{\textbf{\em #1}}

\section{Introduction}
In the past decade, deep learning has achieved remarkable performance in the image classification task.
However, it usually costs a vast number of annotations to train a practical model in a real scenario.
On the contrary, with only a glance at an image of the object even unknown type, humans can quickly and precisely find other same category objects from massive images.
The underlying recognition mechanism of existing deep classifiers is different from humans' recognition mechanism in the classification task.

In fact, humans' quick and precise recognition ability on unknown-type objects benefits from the daily practices on objects of various known categories ~\cite{2010Practice}.
This discovery raises an interesting and vital question: Can the existing deep classifiers quickly and precisely classify novel categories with the support of a set of fully labeled categories?

Some works such as zero/few-shot learning and transfer learning attempt training deep networks to handle novel categories with the help of a set of fully labeled categories.
The former aims to train models using only a few annotated samples,
while the latter focuses on transferring the models learned on one domain to another novel one.
Despite the recent progress  in few-shot and transfer learning,
existing approaches are still prone to either inferior results~\cite{2020Learning},
or the {rigorous requirement that the two tasks are strongly related}~\cite{2020AComprehensive}
and a large number of annotated samples~\cite{2021An,2014Very}.
It seems that the recognition mechanism of the above two kinds of methods still has a difference from the humans' recognition mechanism.
%和已有方法的对比？

When taking an image of the object even unknown type, as a reference, humans can effortlessly find other same category objects from massive images.
Inspired by this fact, we study a new Comparison Knowledge Translation Task (CKT-Task), aiming to \emph{translate} the comparison knowledge learned from massive public source categories where abundant annotations are available, into novel target categories where a few number of annotations or even no annotation are available for each class.
In this paper, comparison knowledge is defined as the recognition ability for distinguishing whether two images belong to the same category.

To this end, we propose a Comparison Classification Translation Network (CCT-Net) for the above CKT-Task.
CCT-Net contains a comparison classifier and a matching discriminator, both of which comprise two branches that take a pair of images as input.
What's more, the matching discriminator has an additional similarity score as input.
The comparison classifier is designed to classify whether two images belong to the same category or not, while the matching discriminator works together in an adversarial manner to ensure whether classified results match the truth.
The comparison classifier only focuses on target categories; meanwhile, the adversarial optimization between the comparison classifier and the matching discriminator will translate the comparison knowledge of source categories into the comparison classifier.

Experiments demonstrate that, with only tens of labeled samples, the proposed CCT-Net achieves close performance on par with fully supervised methods.
When trained with fully annotated samples, CCT-Net achieves state-of-the-art performance.
The most surprising is that the proposed CCT-Net shows promising generalization ability on novel categories.

Our contribution is therefore introducing a new CKT-Task, in aim to translate the comparison knowledge learned from fully annotated source categories into novel ones with few labels,
which emulates the humans’ recognition mechanism in the image classification task.
Furthermore, we propose a dedicated solution CCT-Net that comprises a comparison classifier and a matching discriminator.
 The proposed CCT-Net is evaluated on a broad domain of image datasets, which shows that CCT-Net achieves SOTA classification performance and surprise generalization on novel categories.

 \begin{figure*}[!t]
    \centering
    %\vspace{-0.5em}
    %\includegraphics[scale =0.74]{Figures/framework.pdf}
    \includegraphics[scale =0.75]{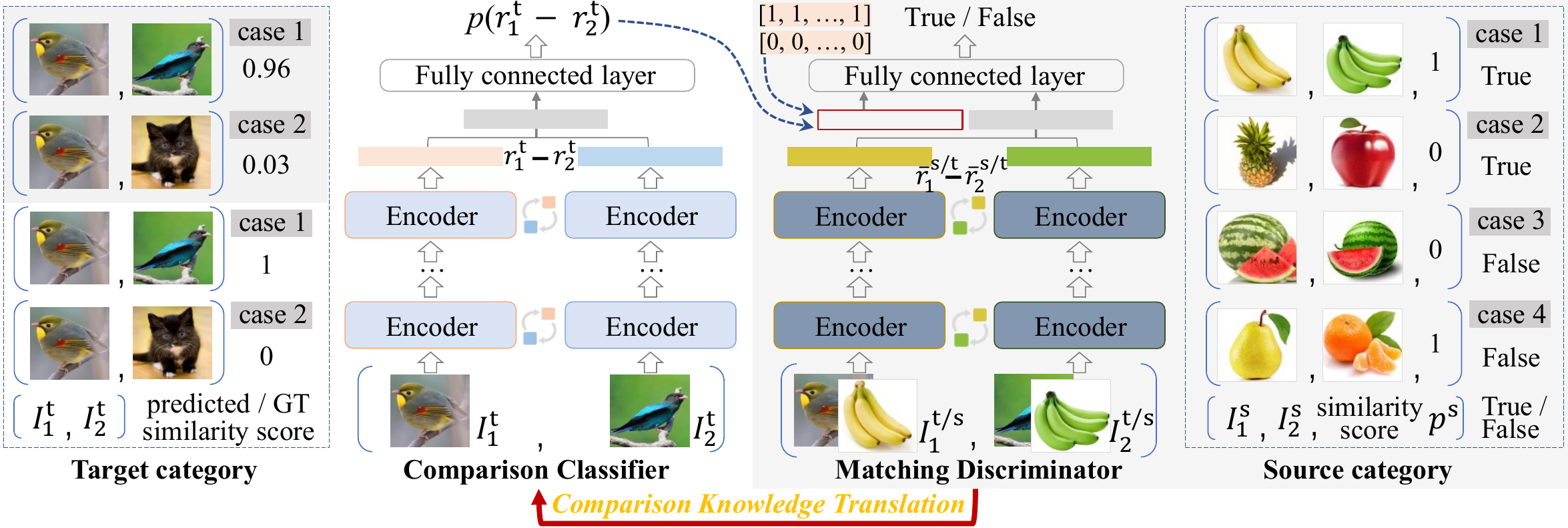}
    \vspace{-1.5em}
    \caption{The framework of CCT-Net composed of a comparison classifier and a matching discriminator.
    The comparison classifier is devised to classify whether the input image pair $(I^t_1,I^t_2)$ belongs to the same category, while the matching discriminator works together in an adversarial manner to ensure whether the predicted result $p(r^t_1-r^t_2)$ matches with the truth. The input of the comparison classifier only contains target categories, which are summarized into two kinds of cases. The input of the matching discriminator is a triplet $(I^{t/s}_1, I^{t/s}_2, p^{t/s})$, which comprises the image pair $(I^{t/s}_1, I^{t/s}_2)$ of target ($t$) and source ($s$) categories with predicted similarity score $p^{t}$ (equals to $p(r^t_1-r^t_2)$) or the assigned similarity score $p^s$. The target category and source category are disjoint.
    }
    \label{fig:framework}
    \vspace{-0.7em}
\end{figure*}

\section{Related Work}
To improve the performance (precision, convergence time, and robustness) of deep classifiers with as few annotations as possible,
various classification tasks, including \emph{zero/few-shot learning}~\cite{2020Learning}, \emph{transfer learning}~\cite{2020AComprehensive}, \emph{distillation learning}~\cite{2020Knowledge}, \emph{un/semi-supervised learning}~\cite{2020A}, have attracted interest from many researchers.
In what follows, we review here two lines of work that are closely related to ours, \emph{zero/few-shot learning} and \emph{transfer learning}.
%In this section, we review here two lines of work that are closely related to ours, \emph{zero/few-shot learning} and \emph{transfer learning}.

\emph{Zero/few-shot learning} can be classified into three categories: mode-based, metric-based, and optimization-based methods. Metric-based methods, including SiameseNet~\cite{2015Siamese}, Match Net~\cite{2016Matching}, Relation Net~\cite{2018Learning}, and Prototype Net~\cite{2017Prototypical} are most related to ours.
SiameseNet~\cite{2015Siamese} is composed of weights shared twin CNNs, which accept a pair of samples as inputs, and their outputs at the top layer are combined in order to output a single pairwise similarity score.
Prototypical Net~\cite{2017Prototypical} classified samples of new categories by comparing the Euclidean distance between the representation of the input sample with a learnable class prototype.
Match Net~\cite{2016Matching} adopted cosine distance to measure similarity between two representations that are encoded with two different encoders.
Unlike Prototypical Net and Matching Net, which use the non-parametric Euclidean distance or cosine distance to measure the similarity between pairwise features,
Relation Net~\cite{2018Learning} adopted a learnable CNN to measure pairwise similarity, which takes the concatenation of feature maps of two samples as input and outputs their relation score.
\cite{2020Learning} summarized more variants about those networks.

Unlike the above methods, we adopt the adversarial manner to distinguish whether the image pair matches with the predicted similarity score.
The major difference is that the comparison knowledge of source categories is translated into the target categories rather than adopting the annotations to supervise the classifier's training.
What's more, the cross-attention mechanism throughout the whole classification process is adopted to enhance the discriminant ability of the comparison classifier.

\emph{Transfer learning} can be categorized into three types: inductive, transductive, and unsupervised transfer learning.
The idea of inductive transfer learning, including multi-task learning algorithm ~\cite{2018Transfer} and self-taught learning ~\cite{2018A}, is to increase approximation of the target probability distribution in the target domain given target tasks are different from the source tasks.
In the transductive transfer learning technique~\cite{2017Annoyed}, the source domain has a lot of labeled data, while the target domain has no labeled data.
Both source tasks and target tasks of transductive transfer learning are similar, whereas there is a difference in the domain only.
Unsupervised transfer learning ~\cite{2018Unsupervised} is the same as inductive , but the main difference is that there is no labeled data in both the source and the target domains.

The differences between transfer learning and CCT-Net contain two aspects: the knowledge type and the way of knowledge transfer.
Transfer learning transfers the discriminant ability from source categories into a classification network for target categories,
while CCT-Net translates comparison classification ability for a pair image, which is a general classification ability for both source and target categories.
For the knowledge transfer way, transfer learning adopts the fine-tune and co-training strategies to transfer the classification ability.
The adversarial training strategy in CCT-Net is adopted to translate the comparison classification ability,
which brings the advantage that redundant discriminant ability for source categories will not distract the discriminant ability for target categories.
%Furthermore, the comparison knowledge brings generalization ability on novel categories than transfer learning techniques.
Furthermore, CCT-Net has superior generalization ability on novel categories, as shown in Section~\ref{section5.3} and Table~\ref{GeneralizationAbility}.

\section{Comparison Knowledge Translation Task}
Inspired by the general boundary knowledge ~\cite{2021Visual,2021Boundary} devised for the segmentation task,
we introduce the comparison knowledge, the recognition ability to distinguish whether two images belong to the same category.
Then, the Comparison Knowledge Translation Task (CKT-Task) is defined as follows.
CKT-Task aims to learn a generalizable image classification framework that can quickly and precisely classify novel categories like humans.
In CKT-Task, there are assumed to be labeled source dataset $\mathbb{S}^m$ that contains $m$ object categories and target dataset $\mathbb{S}^n$ of $n$ object categories.
The $m$ categories and $n$ categories are disjoint.
CKT-Task is supposed to translate the comparison knowledge of $\mathbb{S}^m$ into the comparison classifier $\mathcal{F}_{\theta}$, which is devised for only concentrating on classifying of $n$ categories.
The comparison classifier $\mathcal{F}_{\theta}$ is expected to learn the generalizable distinguishing ability.

\section{Method}
Deep learning methods usually require sufficient annotations to train a well-behaved classifier.
There are vast public classification datasets with plenty of annotations,
which have been exploited by transfer learning and distillation learning for improving classification performance on target domains or categories.
However, the difference specificity of domains and categories severely limits classification performance's upper bound.
In CKT-Task, we study the comparison knowledge that is generalizable for different categories.
To this end, we propose the Comparison Classification Translation Network (CCT-Net) to improve the classification performance on target categories, which draws on the generalizable knowledge from vast public datasets.
Fig.~\ref{fig:framework} depicts the whole framework of CCT-Net comprising a comparison classifier and a matching discriminator.
The adversarial training strategy translates the comparison knowledge of the matching discriminator learned from vast source categories
into the comparison classifier focusing on targeted categories.

\subsection{Comparison Classifier}
In CCT-Net, the comparison classifier $\mathcal{F}_{\theta}$ is designed to be a two-branch architecture. Each branch is composed of multiple encoders.
The comparison classifier only focuses on the dataset $\mathbb{S}^n$ of $n$ target categories.
The input of the comparison classifier is a pair of images, which have two cases similar and non-similar, as shown in the left part in Fig.~\ref{fig:framework}.

With a target image pair $(I^t_1,I^t_2) \in \mathbb{S}^m$ as input, two branches learn a pair of representations $(r^t_1,r^t_2)$.
Then, a fully connected layer predicts the similarity score $p(r^t_1-r^t_2)$ with the representation difference $r^t_1-r^t_2$ as input.

The two branches of the comparison classifier $\mathcal{F}_{\theta}$ share the same architecture but have different parameters.
Inspired by the humans’ recognition mechanism that distinguishes the image pair from global to local,
cross attention is introduced to compare features of two images at each layer.
The comparison classifier will compare basic, middle-level, and high-level semantic features as layers go deeper.

\begin{table*}[!t]
%\scriptsize %\footnotesize% \tiny  %\small \scriptsize \footnotesize
\footnotesize
\centering
\resizebox{\textwidth}{!}{
\begin{tabular}{ccccccccccc}
\toprule
\textbf{ Dataset}   & \multicolumn{2}{c}{\textbf{MNIST}}
    & \multicolumn{2}{c}{\textbf{CIFAR-10}}
    & \multicolumn{2}{c}{\textbf{STL-10}}
    & \multicolumn{2}{c}{\textbf{Oxford-IIIT Pets}}
    & \multicolumn{2}{c}{\textbf{mini-ImageNet}}\\
     \cmidrule(r){2-3}  \cmidrule(r){4-5} \cmidrule(r){6-7} \cmidrule(r){8-9}  \cmidrule(r){10-11}
\textbf{{Method}$\backslash${Index}  }     &Accuracy& F1-score &Accuracy& F1-score   &Accuracy& F1-score &Accuracy& F1-score  &Accuracy& F1-score \\

\cmidrule(r){1-1} \cmidrule(r){2-3}  \cmidrule(r){4-5} \cmidrule(r){6-7} \cmidrule(r){8-9}  \cmidrule(r){10-11}
{\textbf{SCAN}~\cite{2020SCAN}}   &98.2   &98.2  &88.6 &88.6   &77.4  &76.9  &33.6  &25.8   &14.1  &7.2     \\
{\textbf{SimCLR}~\cite{2020A}}   &98.0   &98.0  &95.2 &95.2        &86.0  &86.1  &62.2  &61.4   &44.8  &43.2     \\

\cmidrule(r){1-1} \cmidrule(r){2-3}  \cmidrule(r){4-5} \cmidrule(r){6-7} \cmidrule(r){8-9}  \cmidrule(r){10-11}
{\textbf{Prototype Net}~\cite{2017Prototypical}} &73.5  & 73.5 &66.4  & 65.8  & 75.5  & 73.2  & 51.7 & 50.2 &64.7 & 63.2 \\
{\textbf{Simple CNAPS}~\cite{2020Improved}} &92.6   & 92.6  &74.3  &73.8  & 80.9  & 78.8 & 63.5 & 62.9 &88.4  & 86.9 \\

\cmidrule(r){1-1} \cmidrule(r){2-3}  \cmidrule(r){4-5} \cmidrule(r){6-7} \cmidrule(r){8-9}  \cmidrule(r){10-11}
{\textbf{MixMatch}~\cite{2019MixMatch}}  &98.8   &98.8  &85.4 & 83.9      &90.2   &89.7  &55.0  &53.8  &56.2  &55.2    \\
{\textbf{FixMatch}~\cite{2020FixMatch}}   &99.1   &99.1  &89.7 & 89.2      &92.8  &92.8 &59.5  &58.9  &59.3   &57.2     \\

\cmidrule(r){1-1} \cmidrule(r){2-3}  \cmidrule(r){4-5} \cmidrule(r){6-7} \cmidrule(r){8-9}  \cmidrule(r){10-11}
{\textbf{Transfer}(10)}~\cite{2016Deep}       &91.6  &91.7 &84.8 &84.7  &95.2  &95.2 &89.4 &89.2   &91.7  &91.7    \\
{\textbf{Transfer}(100)}~\cite{2016Deep}        &98.0  &98.0 &92.8 &92.8  &98.4  &98.5  &95.3 &95.3  &95.2  &95.1    \\

\cmidrule(r){1-1} \cmidrule(r){2-3}  \cmidrule(r){4-5} \cmidrule(r){6-7} \cmidrule(r){8-9}  \cmidrule(r){10-11}
{\textbf{SiameseNet}~\cite{2015Siamese}} &99.6  &99.6  &99.3  &99.3  &99.2  &99.2  &97.0  &97.0  & 95.3 & 95.2    \\
{\textbf{VGG-16}~\cite{2014Very}}   &99.6  &99.6 &98.2 &98.2 &98.4  &98.4 &94.0 &94.0&94.5  &94.5    \\
{\textbf{ResNet-50}~\cite{2016Deep}} &\textbf{99.8}  &99.8 &99.0 & 99.0  &99.4  &99.4 &96.8 & 96.4  &96.6  &96.6    \\
{\textbf{MobileNetV2}~\cite{2018MobileNetV2}}   &99.3  &99.3 &98.1 & 98.1 &97.8 &97.8 &95.3 & 95.3   &94.7  &94.7    \\
{\textbf{DenseNet-121}~\cite{2017Densely}}  &99.3  &99.3 &99.0 & 99.0  &99.2  &99.2 &94.6 & 94.6  &96.1  &96.1    \\
{\textbf{ViT-B/16}~\cite{2021An}}   &\textbf{99.8}  &\textbf{99.8} &99.5 & 99.5 &99.4  &99.4 &97.2 & 97.2  &97.7 &97.7    \\

 \cmidrule(r){1-1}    \cmidrule(r){2-3}  \cmidrule(r){4-5} \cmidrule(r){6-7} \cmidrule(r){8-9} \cmidrule(r){10-11}
\textbf{CCT-Net} (0)   &93.6   &93.6  &62.2  &61.4   &80.4   &78.9  &63.9  &64.6 &81.4   &81.1     \\
\textbf{CCT-Net} (20)  &98.0  &98.0 &95.2 &95.2 &98.8  &98.8 &90.2 &90.0   &89.8  &90.0    \\
\textbf{CCT-Net} (100)  &\underline{99.2}  &\underline{99.2} &\underline{97.6} &\underline{97.6} &\underline{99.2}  &\underline{99.2} &\underline{95.7} &\underline{95.7}  &\underline{95.3}  &\underline{95.3}   \\
\textbf{CCT-Net} (all)   &\textbf{99.8} &\textbf{99.8} &\textbf{99.6}  &\textbf{99.6} &\textbf{99.6} &\textbf{99.6}   &\textbf{97.5}  &\textbf{97.5}  &\textbf{97.8}  &\textbf{98.0}    \\
\bottomrule
\end{tabular}
}
\vspace{-0.3em}
\caption{The comparison with SOTA methods. CCT-Net($x$) denotes CCT-Net with $x$ labeled samples per class of the target dataset.
`Bold' and `Underline' indicate the best performance among all methods and all non-fully supervised methods, respectively.
(All scores in $\%$).}
\vspace{-0.8em}
\label{comparing_SOTA}
\end{table*}

\subsection{Matching Discriminator}

As shown in Fig.~\ref{fig:framework}, the matching discriminator has the same architecture as the comparison classifier.
Unlike the comparison classifier, the input of the matching discriminator contains two parts: a pair of images $(I_1,I_2)$ and a similarity score $p \in\{0,1\}$.
The matching discriminator is expected to distinguish whether the input image pair $(I_1,I_2)$ matches with the similarity score $p$.

The input image pairs of the matching discriminator include both the source dataset $\mathbb{S}^m$ and the target dataset $\mathbb{S}^n$.
Given a target image pair $(I^t_1,I^t_2) \in \mathbb{S}^m$ and the similarity score $p(r^t_1-r^t_2)$ predicted by the comparison classifier $\mathcal{F}_{\theta}$,
the matching discriminator first learns a pair of representations $(\bar{r}^t_1,\bar{r}^t_2)$.
Then, the representation difference $\bar{r}^t_1-\bar{r}^t_2$ concatenated with a similarity vector $[p(r^t_1-r^t_2),p(r^t_1-r^t_2),...,p(r^t_1-r^t_2)]$ is input into the fully connected layer of the matching discriminator, which will discriminate whether the similarity of the image pair $(I^t_1,I^t_2)$ matches with the predicted similarity score $p(r^t_1-r^t_2)$.

For the image pair $(I^s_1,I^s_2)$ from the source dataset $\mathbb{S}^m$,
it is assigned with a similarity score $p^s$ (assigning $0$ or $1$ in this paper) and annotated with the matching condition $c^s$ (True or False, set as $1$ or $0$ in the code).
Four kinds of input cases for the source category are summarized in Fig.~\ref{fig:framework}.
Given the source image pair $(I^s_1,I^s_2) \in \mathbb{S}^n$ and the assigned similarity score $p^s$,
the matching discriminator first learn a pair of representations $(\bar{r}^s_1,\bar{r}^s_2)$.
Then, the representation difference $\bar{r}^s_1-\bar{r}^s_2$ concatenated with the similarity vector $[p^s,p^s,...,p^s]$ is input into the fully connected layer of the matching discriminator.
The annotated matching condition $c^s$ for the triplet $(I^s_{1},I^s_{2}, p^s)$ will supervise the matching discriminator $\mathcal{D}_{\phi}$ to learn the matching discrimination ability with the following matching loss function:
\begin{equation}\label{eq1}
\begin{split}
\mathcal{L}_{c}= -c^s\log(\hat{c})+(1-c^s)\log(1-\hat{c}),
\end{split}
\end{equation}
where, $\hat{c}$ denotes the output of the matching discriminator $\mathcal{D}_{\phi}$.

\subsection{Comparison Knowledge Translation}
With the annotations of the source dataset, the matching discriminator can learn the matching discrimination ability.
Then, the adversarial training strategy~\cite{2014Generative} is adopted to translate the comparison knowledge learned by the matching discriminator $\mathcal{D}_{\phi}$ into the comparison classifier $\mathcal{F}_{\theta}$ with the following minimax objective:
\begin{equation}\label{eq2}
\begin{split}
\mathop{\min}_{F_{\theta}} & \mathop{\max}_{D_{\phi}}  \mathop{\mathbb{E}}\limits_{(I^s_1,I^s_2,p^s) \sim \mathbb{P}_{s}}\!\{\log[D_{\phi}(I^s_1,I^s_2,p^s)]\} \\ &+\mathop{\mathbb{E}}\limits_{(I^t_1,I^t_2) \sim \mathbb{P}_{t}}\!\{\log[1-D_{\phi}(I^t_1,I^t_2,\mathcal{F}_{\theta}(I^t_1,I^t_2))]\},
\end{split}
\end{equation}
where $\mathbb{P}_{s}$ and $\mathbb{P}_{t}$ denote the pair data distribution of source and target datasets, respectively.

With the above adversarial training strategy, CCT-Net can be trained in an `unsupervised' learning manner (the target dataset doesn't have any annotation, the source dataset has sufficient annotations).
However, if there are annotations in the target dataset, the binary Cross-Entropy loss will accelerate the training process and improve the final classification performance of the comparison classifier.

Overall, there are three loss functions: the adversarial loss function Eqn.(\ref{eq2}) (for the whole CCT-Net),
the matching loss function Eqn.(\ref{eq1}) (for the matching discriminator), and the binary Cross-Entropy loss function (optional, for comparison classifier).
The complete training algorithm for CCT-Net is summarized in \emph{Algorithm 1\&2 of the supplements}.

\section{Experiments}

\textbf{Dataset.} In the experiments, we adopt five datasets, including MNIST~\cite{1998Gradient}, CIFAR-10~\cite{2009Learning}, STL-10~\cite{2011An}, Oxford-IIIT Pets~\cite{2012Cats}, and mini-ImageNet~\cite{2009ImageNet}, to verify the effectiveness of the proposed CKT-Task and CCT-Net.
The proposed task needs the disjoint source and target categories. So, categories of each dataset are evenly split into the source and target categories in the comparison experiments.

\noindent\textbf{Network architecture and parameter setting.}
For each encoder branch of the comparison classifier and the matching discriminator,
ViT-B/16~\cite{2021An} is adopted as the backbone.
Each encoder of CCT-Net comprises $12$ attention heads, where $2$ attention heads are used for cross-attention between two sub-branches in each layer.
The fully connected layer of the comparison classifier and the matching discriminator share the same architecture (linear layer: $4096$, LeakyReLU, linear layer: $1024$, linear layer: $256$).
The length of the  similarity vector $[p^{t/s},p^{t/s},...,p^{t/s}]$ and the representation difference $r^{t/s}_1-r^{t/s}_2$ is $768$.
More details are given in the \emph{supplements}.

%More details and \emph{source code} are given in the \emph{supplements}.

\begin{table*}[!t]
%\scriptsize %\footnotesize% \tiny  %\small \scriptsize \footnotesize
\footnotesize
\centering
\resizebox{\textwidth}{!}{
\begin{tabular}{ccccccccc}
\toprule
{\textbf{Training$\rightarrow$Validation}}
    & \multicolumn{2}{c}{{\,\,\,\,\,\,\textbf{CIFAR-10 $\unrhd$ STL-10}\,\,\,\,\,\,}}
    & \multicolumn{2}{c}{{\,\,\,\,\,\,\textbf{STL-10 $\unrhd$ CIFAR-10}\,\,\,\,\,\,}}
    & \multicolumn{2}{c}{\textbf{mini-ImageNet $\unrhd$ CIFAR-10}}
    & \multicolumn{2}{c}{{\,\,\textbf{mini-ImageNet $\unrhd$ STL-10}\,\,}}\\
 \cmidrule(r){1-1}    \cmidrule(r){2-3}  \cmidrule(r){4-5} \cmidrule(r){6-7} \cmidrule(r){8-9}
\textbf{Method}$\backslash$\textbf{Index}       &\,\,\,\,\,Accuracy\,\,\,\,\, & F1-score &\,\,\,\,\,Accuracy\,\,\,\,\, & F1-score   &\,\,\,\,\,\,Accuracy\,\,\,\,\,\,& F1-score &\,\,\,\,\,\,Accuracy\,\,\,\,\,\,& F1-score  \\

\cmidrule(r){1-1} \cmidrule(r){2-3}  \cmidrule(r){4-5} \cmidrule(r){6-7} \cmidrule(r){8-9}
\textbf{SiameseNet}(all) &42.4   &34.3  &45.4  &35.0  &42.2   &44.0  &54.4   &51.5 \\

 \cmidrule(r){1-1}    \cmidrule(r){2-3}  \cmidrule(r){4-5} \cmidrule(r){6-7} \cmidrule(r){8-9}

{$\rm{\textbf{CCT}}_{adv.}^{-}(0)$} &24.0   &14.3  &25.4  &17.8   &49.4   &48.2    &79.2   &78.5   \\
{$\rm{\textbf{CCT}}_{adv.}^{-}(20)$}  &79.6   &79.4  &45.6   &45.7   &48.6   &48.9    &85.2   &84.8   \\
{$\rm{\textbf{CCT}}_{adv.}^{-}(all)$}  &89.6   &89.0  &59.2  &60.8   &67.6   &67.3   &95.2   &95.2   \\

 \cmidrule(r){1-1}    \cmidrule(r){2-3}  \cmidrule(r){4-5} \cmidrule(r){6-7} \cmidrule(r){8-9}

{$\rm{\textbf{CCT}}_{cross}^{-}(0)$} &51.6   &48.0  &42.4  &42.6  &23.0  &15.0  &76.8   &76.5 \\
{$\rm{\textbf{CCT}}_{cross}^{-}(20)$} &65.6  &64.6  &51.9   &52.1  &54.6   &53.3  &88.0   &87.3 \\
{$\rm{\textbf{CCT}}_{cross}^{-}(all)$} &87.6   &87.0  &56.2   &57.2  &62.4   &62.5  &90.4  &90.3\\

 \cmidrule(r){1-1}    \cmidrule(r){2-3}  \cmidrule(r){4-5} \cmidrule(r){6-7} \cmidrule(r){8-9}
\textbf{CCT-Net}(0)  &56.0  &55.8   &53.2  &51.4  &54.4   &52.8  &79.2   &78.7     \\
\textbf{CCT-Net}(20) &86.0 &86.1   &60.0 &59.1   &67.4  &67.4  &95.2 &95.2   \\
\textbf{CCT-Net}(all) &90.0 & 89.7  &62.0 & 63.0    &70.4  &70.2  &98.8  &98.8   \\
\bottomrule
\end{tabular}}
\vspace{-0.3em}
\caption{Generalization results on novel categories.
`dataset1 $\unrhd$ dataset2' denotes CCT-Net is only trained on dataset1 and tested on unseen dataset2.
$\mathbi{CCT}_{adv.}^{-}$ and $\mathbi{CCT}_{cross}^{-}$
denote CCT-Net without discriminator and cross attention, respectively.}
\vspace{-0.5em}
\label{GeneralizationAbility}
\end{table*}

\subsection{Comparing with SOTA Methods}
In this section, the proposed method is compared with \emph{unsupervised methods}: SCAN~\cite{2020SCAN} and SimCLR~\cite{2020A},
\emph{few-shot methods}: Prototype Net~\cite{2017Prototypical} and Simple CNAPS~\cite{2020Improved},
\emph{semi-supervised methods}: MixMatch~\cite{2019MixMatch} and FixMatch~\cite{2020FixMatch},
\emph{transfer learning methods}: Transfer(10) and Transfer(100),
and \emph{fully supervised methods}: SiameseNet~\cite{2015Siamese}, VGG-16~\cite{2014Very}, ResNet-50~\cite{2016Deep}, MobileNetV2~\cite{2018MobileNetV2}, DenseNet-121~\cite{2017Densely} and
ViT-B/16 [Dosovitskiy \emph{et al.}, 2021].

The fully and semi-supervised methods are only trained on the target categories in the experiment.
Transfer(10) denotes the ResNet-50 trained with $10$ annotated samples of the target category.
$80\%$ of the target datasets are used as annotated samples for semi-supervised methods (MixMatch and FixMatch).
Five annotated samples of target categories are used for the few-shot methods (Prototype Net and Simple CNAPS).
Table~\ref{comparing_SOTA} shows the quantitative comparison results,
where we can see that CCT-Net with fully annotated samples achieves the SOTA performance on par with all existing methods.
With $100$ annotated pairs, CCT-Net achieves the best performance among all non-fully supervised methods.
It's noted that CCT-Net achieves higher classification performance than few-shot methods and semi-supervised methods even without annotated samples of target category,
which demonstrates the effectiveness of the proposed comparison knowledge translation.

\begin{table}[!t]
%\scriptsize %\footnotesize% \tiny  %\small \scriptsize \footnotesize
\footnotesize
\centering
\resizebox{0.48\textwidth}{!}{
\begin{tabular}{llcc}
\toprule
\textbf{Source}$\rightarrow$\textbf{Target} & $N_s$$\rightarrow$$N_t$ &  Accuracy& F1-score\\
\midrule
STL-10$\rightarrow$MNIST &10$\rightarrow$10 & 90.0 &90.0  \\
Single$\rightarrow$Single   & 1$\rightarrow$1 &  100.0  &100.0   \\
CIFAR-10$\rightarrow$STL-10 & 10$\rightarrow$10 &89.0 & 89.1 \\
STL-10$\rightarrow$CIFAR-10 & 10$\rightarrow$10 & 79.4 & 79.4 \\
mini-ImageNet$\rightarrow$CIFAR-10   &100$\rightarrow$10 &  73.6 & 73.8 \\
mini-ImageNet$\rightarrow$STL-10     &100$\rightarrow$10 &  90.2 &90.1  \\

\bottomrule
\end{tabular}}
\vspace{-0.3em}
\caption{The translation results between different dataset settings.
 '{Source}$\rightarrow${Target}' denotes translating knowledge of the source dataset into the comparison classifier for the target dataset.
  $N_s$ and $N_t$ denote category number of source dataset and target dataset, respectively.
  'Single$\rightarrow$Single' denotes translating knowledge of ten categories, each of which is randomly selected from mini-ImageNet at each time, into the lion category.}
\vspace{-0.5em}
\label{task_on_datasets}
\end{table}

\subsection{Translation between Different Dataset}

This section provides the knowledge translation experiments between different datasets to verify the robustness of CKT-Task and CCT-Net.
For all the translation tasks (Source$\rightarrow$Target) in Table~\ref{task_on_datasets}, only $20$ annotated pairs of the target categories are used for CCT-Net.
In the experiment, different datasets and different numbers of source and target categories are taken as two ablation factors.
From Table~\ref{task_on_datasets}, we can see that all the translation tasks achieve satisfactory results between different datasets with different source and target category numbers,
which verifies the high extensibility and practicability of CKT-Task and CCT-Net.

\subsection{Generalization Ability on Novel Category}
\label{section5.3}
To further verify the generalization ability of CKT-Task, the trained models are directly tested on novel categories (the trained models have never seen before).
Table~\ref{GeneralizationAbility} shows the generalization results on three datasets.
`dataset1 $\unrhd$ dataset2' denotes CCT-Net is only trained on the dataset1 ( all categories of dataset1 are evenly split into the source and target categories)
and then tested on unseen dataset2 directly.
Due to the specific network architecture, most existing classification methods can't be directly tested on a novel category.
So, we only compare the proposed method with SiameseNet, which has the same two-branch architecture.
`SiameseNet(all)' denotes that SiameseNet is trained with fully annotated samples of dataset1 and then tested on dataset2.

From Table~\ref{GeneralizationAbility}, we can see that CCT-Net(all) achieves about double increment than SiameseNet(all) on all datasets
 ( CIFAR-10 $\unrhd$ STL-10: $(+47.6,+55.4)$, STL-10 $\unrhd$ CIFAR-10: $(+16.6,+28.0)$,  mini-ImageNet $\unrhd$ CIFAR-10: $(+28.2,+26.2)$,  mini-ImageNet $\unrhd$ STL-10: $(+44.4, +47.3)$),
 which demonstrates the surprising generalization ability of CKT-Task and CCT-Net on novel categories.
Even without annotations of target categories, CCT-Net(0) still achieves better generalization ability than SiameseNet(all).
What's more, we further verify the importance of each component of CCT-Net on generalization ability.
With annotated samples of target categories,
$\mathbi{CCT}_{cross}^{-}$ (CCT-Net without cross attention) achieves lower classification accuracy than $\mathbi{CCT}_{adv.}^{-}$ (CCT-Net without discriminator), which demonstrates that cross attention mechanism  contributes more to the generalization ability of CCT-Net.

\subsection{Convergence Speed on Novel Category}
\begin{figure}[!t]
    \centering
    %\vspace{-0.5em}
    %\includegraphics[scale =0.74]{Figures/framework.pdf}
    \includegraphics[scale =0.28]{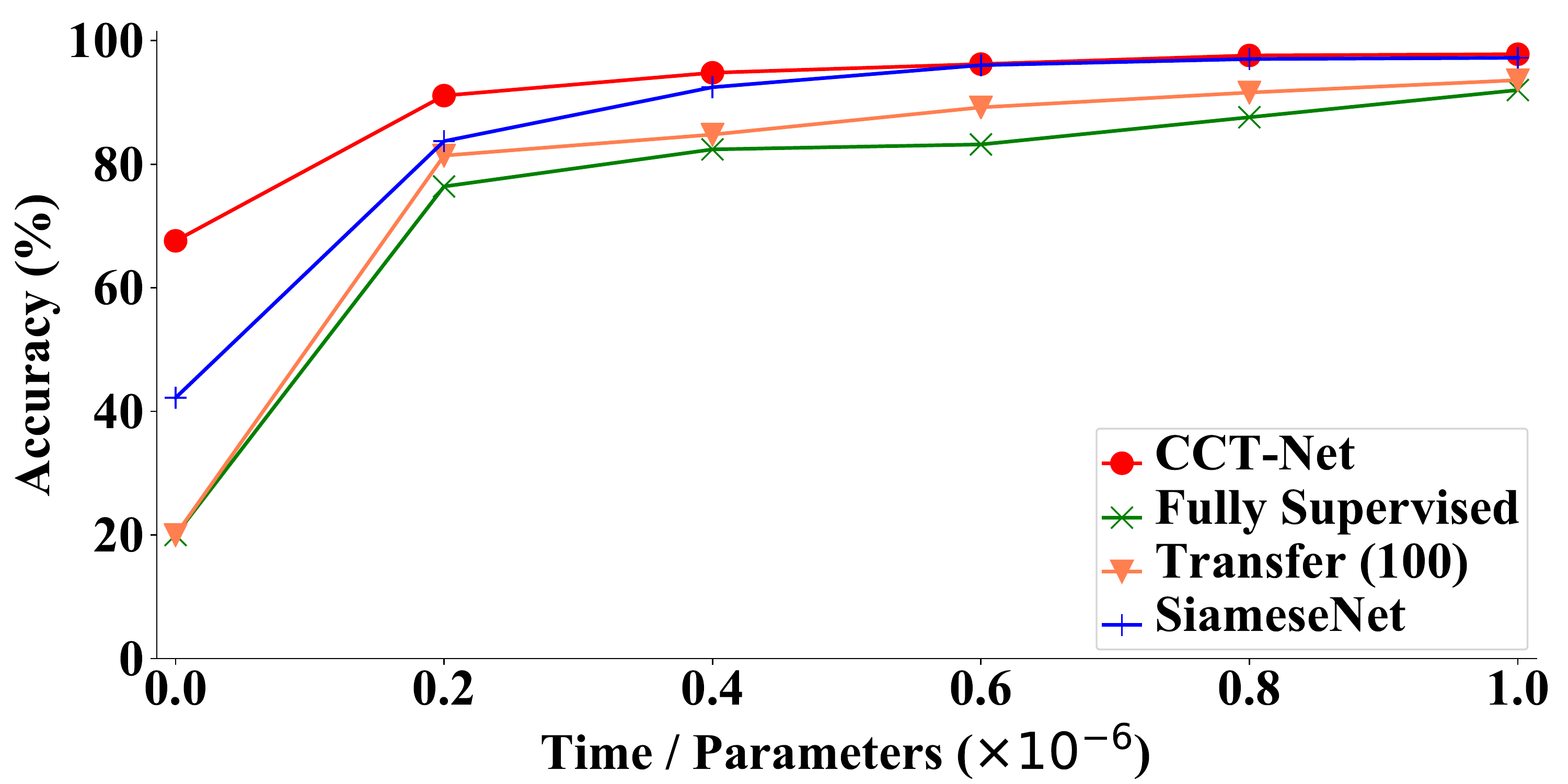}
    \caption{The convergence speed comparison of different methods.}
    %\vspace{-0.5em}
    \label{convergenceComparison}
\end{figure}

Except for the excellent generalization ability, CKT-Task and CCT-Net still have the advantage of fast convergence speed, shown in Fig.~\ref{convergenceComparison}.
Due to the different number of model parameters, we adopt  `Time / parameters' (training time (seconds) per 1M parameter)
to compare the convergence speed of different methods on novel categories.
From Fig.~\ref{convergenceComparison},
we can see that the proposed CCT-Net have a faster convergence speed on novel category than existing methods,
which indicates that CCT-Net imitates the humans' quick recognition ability on novel category.

\subsection{Accuracy Trend with Incremental Category}
\begin{figure}[!t]
    \centering
    %\vspace{-0.5em}
    %\includegraphics[scale =0.74]{Figures/framework.pdf}
    \includegraphics[scale =0.36]{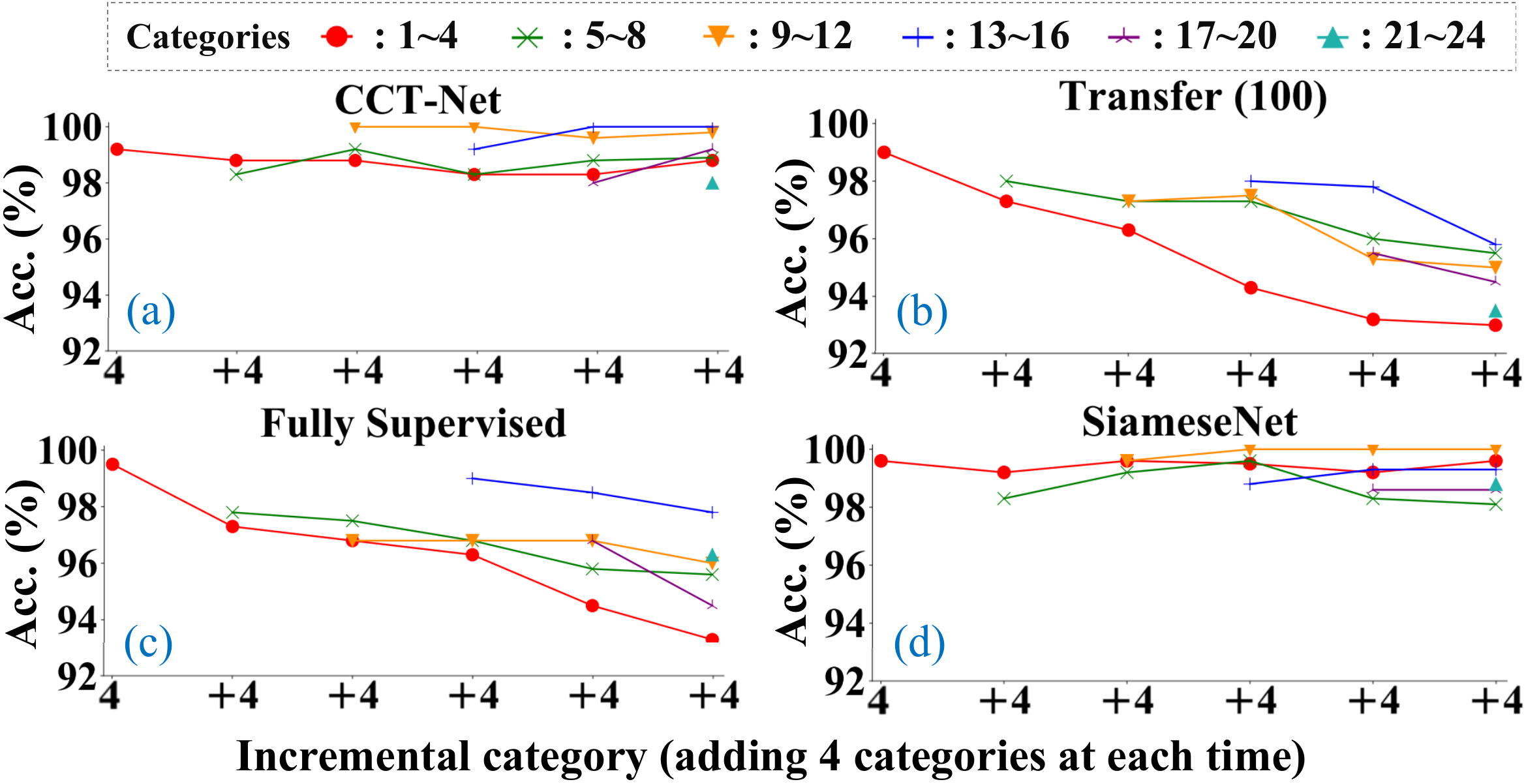}
    \vspace{-1.5em}
    \caption{The accuracy trend with incremental category.}
    %\vspace{-0.5em}
    \label{AccuracyTrend}
\end{figure}

Fig.~\ref{AccuracyTrend} shows the classification accuracy trend of different methods
with incremental category on mini-ImageNet.
The experiments are validated on $28$ randomly selected categories.
In the whole training process, each method is firstly trained with $4$ categories.
Next, the previous model is trained on cumulative categories (adding $4$ categories at each time) when the previous model converges.

From Fig.~\ref{AccuracyTrend}(b)(c),
we can see that the fully supervised method and the transfer learning method have descending trends of classification accuracy with incremental category.
The fundamental reason is that the learned classification knowledge of the fully supervised method and the transfer learning method is highly associated with the category.
When the category number increases, the recognition ability of the deep model for each category will decrease.
On the contrary, the proposed CCT-Net and SiameseNet achieve an ascending trend of classification accuracy with incremental category, which verifies that the learned comparison knowledge is generalizable for the classification task. What's more, in the last three increment phases,  all categories for CCT-Net achieve ascending trend (Fig.~\ref{AccuracyTrend}(a)), while
`categories 5$\sim$8' for SiameseNet achieves descending trend (Fig.~\ref{AccuracyTrend}(d)).

\begin{table}[!t]
%\scriptsize  %
\footnotesize %\footnotesize% \tiny  %\small \scriptsize
\centering
\resizebox{0.48\textwidth}{!}{
\begin{tabular}{cccccccc}
\toprule
\textbf{Index}$\backslash$\textbf{Annotations}
& 0
& 5
& 10
& 20
& 50
& 100
& all \\
\cmidrule(r){1-1} \cmidrule(r){2-8}
Accuracy       &81.4    &89.0 & 89.5    &89.8  &94.9 &95.3  &97.8   \\
%\cmidrule(r){1-1}     \cmidrule(r){2-2}  \cmidrule(r){3-3}   \cmidrule(r){4-4}  \cmidrule(r){5-5}
F1-score   &81.1   &89.2 & 89.5  &90.0  &94.7 &95.3  &98.0   \\
%\cmidrule(r){1-1}     \cmidrule(r){2-2}  \cmidrule(r){3-3}   \cmidrule(r){4-4}  \cmidrule(r){5-5}
\bottomrule
\end{tabular}
}
%\vspace{-0.3em}
\caption{The ablation study on annotated samples of CCT-Net.}
\label{ablation_annotation}
\vspace{-0.5em}
\end{table}

\subsection{Ablation Study}

To verify the effectiveness of CCT-Net's components, we do ablation study on the false pair samples (case 3 and case 4 in Fig.~\ref{fig:framework}), discriminator, cross attention, input position of the condition $c^{t/s}$, and the different numbers of labeled samples of target categories.
For the ablation study, the mini-ImageNet dataset is adopted.
For $CCT_{false}^{-}$, $CCT_{adv.}^{-}$, $CCT_{cross}^{-}$ and $CCT_{head}^{cond.}$,
$20$ labeled pairs for each target category are used. From Table~\ref{ablationComponent}, we can see that CCT-Net(20) achieves higher scores than other ablative methods, demonstrating all components' effectiveness.
$CCT_{adv.}^{-}$ achieves the lowest score than other ablative methods, which indicates that the adversarial translation strategy is the most critical factor for CKT-Task and CCT-Net.

The ablation study results on annotated pairs of the target category are given in Table~\ref{ablation_annotation}, where we can find that $50$-annotated-pairs is a critical cut-off point, supplying relatively sufficient guidance. Even without any guidance of labeled pairs of target categories,
CCT-Net still achieves $81.4\%$ accuracy score on target categories, verifying the effectiveness of CKT-Task and CCT-Net again.

\begin{table}[!t]
%\scriptsize  %
\footnotesize %\footnotesize% \tiny  %\small \scriptsize
\centering
\resizebox{0.48\textwidth}{!}{
\begin{tabular}{cccccc}
\toprule
\textbf{Index}$\backslash$\textbf{Ablation}
& $\mathbi{CCT}_{false}^{-}$
& $\mathbi{CCT}_{adv.}^{-}$
& $\mathbi{CCT}_{cross}^{-}$
& $\mathbi{CCT}_{head}^{cond.}$
& \textbf{CCT-Net} (20)\\
\cmidrule(r){1-1} \cmidrule(r){2-6}
Accuracy      &74.9  &67.8 &89.0 &71.3&  89.8 \\
%\cmidrule(r){1-1}     \cmidrule(r){2-2}  \cmidrule(r){3-3}   \cmidrule(r){4-4}  \cmidrule(r){5-5}
F1-score  &76.0 &69.5 &89.2 &72.6 & 90.0\\
%\cmidrule(r){1-1}     \cmidrule(r){2-2}  \cmidrule(r){3-3}   \cmidrule(r){4-4}  \cmidrule(r){5-5}
\bottomrule
\end{tabular}
}
%\vspace{-0.3em}
\caption{The ablation study results of CCT-Net.
$\mathbi{CCT}_{false}^{-}$, $\mathbi{CCT}_{adv.}^{-}$, $\mathbi{CCT}_{cross}^{-}$
denote CCT-Net without false pair samples, discriminator, cross attention, respectively. $\mathbi{CCT}_{head}^{cond.}$ denotes placing the similarity score in the head of the discriminator.}
\label{ablationComponent}
%\vspace{-0.5em}
\end{table}

\section{Conclusion}

This paper studies a new Comparison Knowledge Translation Task (CKT-Task), which imitates humans' quick and precise recognition ability on novel categories.
The goal of CKT-Task is to translate the comparison knowledge of source categories into the deep classifier for new categories in the least effort and dependable way.
Toward realizing CKT-Task, we introduce the Comparison Classification Translation Network (CCT-Net), which comprises a comparison classifier and a matching discriminator.
The comparison classifier is devised to classify whether two images belong to the same category or not, while the matching discriminator works together in an adversarial manner to ensure whether classified results match the truth.
With the adversarial training between the comparison classifier and the matching discriminator, the comparison knowledge of massive public source categories is successfully translated into the deep classifier for target categories.
There is no special requirement for the source and target categories, which means that all public classification datasets can be used as the source datasets in the proposed CCT-Net.
Exhaustive experiments show that CCT-Net achieves impressive generalization ability and SOTA performance on par with existing methods.
Surprisingly, CCT-Net also achieves impressive results, including fast convergence speed and high accuracy on novel categories, revealing its superior generalization ability.
We will focus on exploring more generalization knowledge and framework on other tasks in the future.

\section*{Acknowledgments}
This work is supported by National Natural Science Foundation of China (No.62002318),Key Research and Development Program of Zhejiang Province (2020C01023), Zhejiang Provincial Science and Technology Project for Public Welfare (LGF21F020020), Fundamental Research Funds for the Central Universities (2021FZZX001-23),  Starry Night Science Fund of Zhejiang University Shanghai Institute for Advanced Study (Grant No. SN-ZJU-SIAS-001), and Zhejiang Lab (No.2019KD0AD01/014).

\section*{Supplementary Material}

In the supplementary materials, we provide experiment details, the unsupervised and supervised training algorithms, and the analysis of the similarity score $p^s$.
%For the code of those experiments, please refer to the ``CodeAppendix'' file.

\begin{algorithm}[!h]
\caption{Unsupervised Training Algorithm for CCT-Net}
\begin{algorithmic}[1]
\renewcommand{\algorithmicrequire}{\textbf{Require:}}
\renewcommand{\algorithmicensure}{\textbf{Require:}}
\Require{ The interval iteration number $n_{dis}$, the batch size $K$, Adam hyperparameters $\alpha,\beta_{1},\beta_{2}$, the balance parameters $\eta$ for $\mathcal{L}_{c}$.}
\Require{ Initial matching discriminator parameters $\phi$, and comparison classifier parameters $\theta$. }
\While{$\theta$ has not converged}
\For{$t=1,...,n_{dis}$}
\For{$k=1,...,K$}
\State Sample image pair $(I^t_1,I^t_2)$ from target dataset $\mathbb{S}^m$.
\State Sample true triplet $\{(I^s_1,I^s_2,q^s_{true}),c^s_{true}\}$ (case 1\&2) and false triplet $\{(I^s_1,I^s_2,q^s_{false}),c^s_{false}\}$ (case 3\&4) from source dataset $\mathbb{S}^n$.
\State $\mathbf{\mathcal{L}_{dis}}^{(k)} \leftarrow \log[D_{\phi}(I^s_1,I^s_2,p^s_{true})]$+$\log[1-D_{\phi}(I^t_1,I^t_2,\mathcal{F}_{\theta}(I^t_1,I^t_2))]$.
\State $\hat{c}_{true}=D_{\phi}(I^t_1,I^t_2,q^s_{true})$.
\State $\hat{c}_{false}=D_{\phi}(I^t_1,I^t_2,q^s_{false})$.
\State $\mathbf{\mathcal{L}_{c}}^{(k)} \leftarrow -c^s_{true}\log(\hat{c}_{true})-(1-c^s_{true})\log(1-\hat{c}_{true})-c^s_{false}\log(\hat{c}_{false})-(1-c^s_{false})\log(1-\hat{c}_{false})$.
\EndFor
\State $\phi \leftarrow Adam(\nabla_{\phi}\frac{1}{K} \sum^K_{k=1}(\mathbf{\mathcal{L}_{dis}}^{(k)}+\eta \mathbf{\mathcal{L}_{c}}^{(k)}),\phi,\alpha,\beta_1,\beta_2)$.
\EndFor
\State Sample unlabeled batch  $\{(I^t_1,I^t_2)^{(k)}\}^K_{k=1}$  from target dataset $\mathbb{S}^m$.
\State  $\mathbf{\mathcal{L}_{cls}}^{(k)} \leftarrow  \log[\mathcal{D}_{\phi}(I^t_1,I^t_2,\mathcal{F}_{\theta}(I^t_1,I^t_2))]$.
\State $\theta \leftarrow Adam(\nabla_{\theta}\frac{1}{K} \sum^K_{k=1}-\mathbf{\mathcal{L}_{cls}}^{(k)},\theta,\alpha,\beta_1,\beta_2)$.
\EndWhile
\State \Return Comparison classifier parameters $\theta$, matching discriminator parameters $\phi$.
\end{algorithmic}
\label{alg:alg1}
\end{algorithm}

%%%%%%%%% BODY TEXT
\section*{A. Experiment Parameters Setting}

The parameters for training CCT-Net are set as follows: the batch size $K =32$, the  interval iteration number $n_{dis}=3$, Adam hyperparameters $\alpha =0.001,\beta_{1}=0.9,\beta_{2}=0.999$. The learning rates for the comparison classifier and the matching discriminator are set to be $2e^{-6}$ and $1e^{-4}$ with cosine annealing.

\begin{table}[!t]
%\scriptsize %\footnotesize% \tiny  %\small \scriptsize \footnotesize
\footnotesize
\centering
\resizebox{0.45\textwidth}{!}{
\begin{tabular}{ccccc}
\toprule
\textbf{{Method} }  & \multicolumn{2}{c}{\textbf{CCT-Net} (100)  }
    & \multicolumn{2}{c}{\textbf{CCT-Net} (all) }\\
     \cmidrule(r){2-3}  \cmidrule(r){4-5}
 \textbf{ Value of $p^s$ $\backslash $ {Index}}      &Acc.& F1-score &Acc.& F1-score   \\

$[1]$ \& $[0]$            &95.3  & 95.3 &97.8 & 98.0 \\
 $[0.99,1]$ \& $[0,0.01]$  &94.7  & 94.7 & 98.3 &98.1 \\
$[0.9,1]$ \& $[0,0.1]$   &95.2  & 95.2  & 97.9 &97.4  \\
$[0.8,1]$ \& $[0,0.2]$    &94.6  & 94.6  & 98.5 &97.4\\
$[0.7,1]$ \& $[0,0.3]$  &95.1  & 95.1 &98.3 & 97.9 \\
\bottomrule
\end{tabular}
}
\vspace{-0.3em}
\caption{The ablation study on the value of the similarity score $p^s$.
`$[0]$ \& $[1]$' denotes that values of $p^s$ are set to $0$ and $1$ when the two images belong to the same category or not, respectively.
`$[v_1,1]$ \& $[0,v_2]$' denotes that $p^s$ is set to the value randomly sampled from $[v_1,1]$ and $[0,v_2]$ when the two images belong to the same category or not, respectively.
(All scores in $\%$).}
%\vspace{-0.8em}
\label{ablationOnValue}
\end{table}

\begin{algorithm}[!t]
\caption{Supervised Training Algorithm for CCT-Net}
\begin{algorithmic}[1]
\renewcommand{\algorithmicrequire}{\textbf{Require:}}
\renewcommand{\algorithmicensure}{\textbf{Require:}}
\Require{ The interval iteration number $n_{dis}$, the batch size $K$, Adam hyperparameters $\alpha,\beta_{1},\beta_{2}$, the balance parameters $\eta,\tau$ for $\mathcal{L}_{c}$ and $\mathbf{\mathcal{L}_{sup}}$.}
\Require{ Initial matching discriminator parameters $\phi$, and comparison classifier parameters $\theta$. }
\While{$\theta$ has not converged}
\For{$t=1,...,n_{dis}$}
\For{$k=1,...,K$}
\State Sample image pair $(I^t_1,I^t_2)$ from target dataset $\mathbb{S}^m$.
\State Sample true triplet $\{(I^s_1,I^s_2,q^s_{true}),c^s_{true}\}$ (case 1\&2) and false triplet $\{(I^s_1,I^s_2,q^s_{false}),c^s_{false}\}$ (case 3\&4) from source dataset $\mathbb{S}^n$.
\State $\mathbf{\mathcal{L}_{dis}}^{(k)} \leftarrow \log[D_{\phi}(I^s_1,I^s_2,p^s_{true})]$+$\log[1-D_{\phi}(I^t_1,I^t_2,\mathcal{F}_{\theta}(I^t_1,I^t_2))]$.
\State $\hat{c}_{true}=D_{\phi}(I^t_1,I^t_2,q^s_{true})$.
\State $\hat{c}_{false}=D_{\phi}(I^t_1,I^t_2,q^s_{false})$.
\State $\mathbf{\mathcal{L}_{c}}^{(k)} \leftarrow -c^s_{true}\log(\hat{c}_{true})-(1-c^s_{true})\log(1-\hat{c}_{true})-c^s_{false}\log(\hat{c}_{false})-(1-c^s_{false})\log(1-\hat{c}_{false})$.
\EndFor
\State $\phi \leftarrow Adam(\nabla_{\phi}\frac{1}{K} \sum^K_{k=1}(\mathbf{\mathcal{L}_{dis}}^{(k)}+\eta \mathbf{\mathcal{L}_{c}}^{(k)}),\phi,\alpha,\beta_1,\beta_2)$.
\EndFor
\State Sample labeled batch $\{(\check{I}^t_1,\check{I}^t_2)^{(k)},q^t\}^K_{k=1}$ and unlabeled batch  $\{(I^t_1,I^t_2)^{(k)}\}^K_{k=1}$  from target dataset $\mathbb{S}^m$.
\State  $\mathbf{\mathcal{L}_{sup}}^{(k)}\leftarrow CrossEntropy(\mathcal{F}_{\theta}(\check{I}^t_1,\check{I}^t_2)^{(k)},q^t)$.
\State  $\mathbf{\mathcal{L}_{cls}}^{(k)} \leftarrow  \log[\mathcal{D}_{\phi}(I^t_1,I^t_2,\mathcal{F}_{\theta}(I^t_1,I^t_2))]$.
\State $\theta \leftarrow Adam(\nabla_{\theta}\frac{1}{K} \sum^K_{k=1}\{\tau \mathbf{\mathcal{L}_{sup}}^{(k)}-\mathbf{\mathcal{L}_{cls}}^{(k)}\}),\theta,\alpha,\beta_1,\beta_2)$.
\EndWhile
\State \Return Comparison classifier parameters $\theta$, matching discriminator parameters $\phi$.
\end{algorithmic}
\label{alg:alg2}
\end{algorithm}

\section*{B. Training Algorithm for CCT-Net}

As described in the main body of the manuscript, the proposed CCT-Net can be trained in an unsupervised  or supervised manner.
When there are annotated pairs for the target dataset, CCT-Net will be trained with Algorithm~\ref{alg:alg1} in an unsupervised manner.
On the other hand, the annotated pairs of the target dataset can accelerate the training process and improve the final classification performance of the comparison classifier.
So, we summarize the supervised training algorithm for CCT-Net, shown in Algorithm~\ref{alg:alg2}.

\section*{C. Discussion of Annotated Similarity Score $p^s$ }
The input of the matching discriminator contains both the triplet of source categories and target categories.
For the target categories, the value of the predicted similarity score $p^t$ is usually between $0$ and $1$.
Due to the characteristics of the log function, $p^t$ will never equal to $0$ and $1$.
How to set the value of similarity score $p^s$ for the image pair of source categories?

We conducted an ablation study on the value of $p^s$. There are two kinds of situations (fixed value `$[0]$ \& $[1]$', random value sampled from `$[v_1,1]$ \& $[0,v_2]$') for the annotated similarity score $p^s$.
Table~\ref{ablationOnValue} shows the classification performance with different values of $p^s$,
 where we can see that the classification performance has small numerical fluctuations,
 which indicates that the different values of similarity score $p^s$ have the same effect for the classification of CCT-Net.
 So, in the experiment of manuscript, we adopt the setting of `$[0]$ \& $[1]$', which means that the values of $p^s$ are set to $0$ and $1$ when the two images of a pair belong to the same category or not, respectively.
For the condition value, \cite{2016Semantic} also  gave similar experiments results as ours.

\bibliographystyle{named}
\bibliography{ijcai22}

% 附加材料里，数据集的划分，以及 算法 流程图  In the training stage, CCT-Net(20) and CCT-Net(20) are firstly pretrained on

%\newpage

\end{document}